# Tucker Bilinear Attention Network for Multi-scale Remote Sensing Object Detection

Tao Chen , *Student Member, IEEE*, Ruirui Li*, *Member, IEEE*, Jiafeng Fu, and Daguang Jiang

*Abstract*—Object detection on very high resolution (VHR) remote sensing images is crucial for applications such as urban planning, land resource management, and rescue missions. However, the large-scale variation of remote-sensing targets presents a significant challenge for VHR remote-sensing object detection. Although existing methods have improved the accuracy of high-resolution remote sensing object detection through feature pyramid structure enhancement and various attention modules, small targets still suffer from missed detections due to the loss of key detail features. There is still room for improvement in terms of multiscale feature fusion and balance. To address this issue, this paper proposes two novel modules: Guided Attention and Tucker Bilinear Attention, which are applied to the stages of early fusion and late fusion, respectively. The former effectively retains critical detail features, while the latter balances features through semantic-level correlation mining. Based on these two modules, a new multiscale remote sensing object detection framework is presented. This method significantly improves the average precision of small objects and achieves the highest mean average precision when compared with nine state-of-the-art methods on DOTA, DIOR, and NWPU VHR-10 datasets.Code and models are available at https://github.com/Shinichict/GTNet.

## I. INTRODUCTION

Object detection is an intricate process that combines two important tasks, namely localization and classification. Localization refers to the process of drawing bounding boxes around objects, while classification involves the identification of different classes. Advanced detection networks currently use both one-stage and two-stage models. Some of the popular one-stage models include RetinaNet[1], CenterNet[2], and YOLO-v5[3], while two-stage models include Faster R-CNN[4], Cascade R-CNN[5], and Libra R-CNN[6]. These models have performed remarkably well on general tasks involving natural image datasets. However, remote sensing image processing presents more challenges due to variations in object scale and orientation. In remote sensing images, large targets such as airports and ports exist alongside small and densely packed targets such as cars and ships. Moreover, the size of objects in remote sensing images varies from 800*800 to 16*16 pixels, causing a decrease in average precision (AP) as the size of the object decreases. This is mainly attributed to the network's limited attention to key detailed features. Methods specific to remote sensing object detection have been further developed and discussed in detail in [18].

Multi-scale feature extraction techniques are widely used to improve the perception of detailed information. The Feature Pyramid Network (FPN), introduced in [7], is an example of such a technique that generates multi-layer feature maps by integrating deep and shallow features for object detection. Improved versions of FPN[8] are extensively used in multi-scale remote sensing object detection. FSoD-Net [9] has recently proposed a single-stage full-scale object detection network that incorporates a multi-scale feature enhancement mechanism. Despite preserving shallow features to a certain extent for subsequent learning, these techniques still encounter difficulties in efficiently detecting small objects. As large and small objects' detailed information exist concurrently, the saliency of large objects often dominates, leading to a weakening of the features of small objects, thereby hindering their effective detection.

Contextual information is crucial in visual recognition tasks and attention mechanisms represent a commonly used technique to grasp it. MA-FPN proposes multi-attention feature pyramid network, whereas CANet [10] leverages cross-layer attention network, which merges deep and shallow layers through bidirectional feature fusion.Typically, these attention modules are independently implemented at corresponding levels and seldom delve into the contextual correlation between layers. CANet, despite finding contextual correlation helpful, only integrates the extracted attention features, implying a failure to comprehensively explore the feature associations across layers.

Bilinear pooling is a feature fusion method that enables comprehensive recognition of feature associations by leveraging outer product and average pooling techniques to consolidate features from variant networks into a bilinear feature representation. However, the calculation process incurs quadratic dimension growth, limiting its potential application in terms of scope and scale. For instance, high-dimensional input features and excessive use of bilinear pooling are avoided. To alleviate computational cost, some low-rank bilinear pooling methods capitalize on the sparsity of feature relationships. Inspired by these established techniques, our study aims to extensively scrutinize and exploit feature correlations amid layers through implementing a low-rank bilinear pooling module, thereby optimizing the detection accuracy of full-scale remote sensing objects.

This paper proposes two cross-layer feature fusion modules for the early-stage fusion and the late-stage fusion, which are: Guided Attention(GA) and Tucker Bilinear Attention(TBA) respectively. The former attempts to guide and enhance low-level features with higher-level semantics.

This work was supported by the National Natural Science Foundation of China under Grant 62101021. (*Corresponding author: Ruirui Li.*)

Tao Chen, Daguang Jiang, Jiafeng Fu, and Ruirui Li are with the College of Information Science and Technology, Beijing University of Chemical Technology, Beijing, 100029, China. (e-mail: zero2zerox@163.com, 1994500057@buct.edu.cn, 1138823978@qq.com, ilydouble@gmail.com,).

The latter decomposes the tensor obtained by bilinear transformation with Tucker factorization so that the correlation of cross-layer features can be calculated effectively through end-to-end learning. The two proposed modules can be used as more basic functional units to be embedded in two-stage detection frameworks.

Below, we summarize the main contributions of this work:
- We design the Guided Attention(GA) module which utilizes deeper features to guide shallower features to better preserve potentially useful details.
- We present Tucker Bilinear Attention(TBA) Network to learn the correlations of cross-layer features output by FPN, so as to better fuse them to improve multi-scale object detection accuracy.
- Based on GA and TBA, this paper constructs a remote-sensing object detection framework and conducts extensive experiments on three public datasets.

## II. PROPOSED METHOD

### A  Overall Architecture

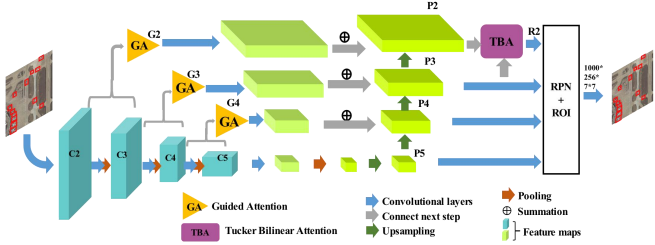

Fig. 1. The proposed framework for remote sensing multi-scale object detection

The two-stage detection framework proposed in this paper employs the Cascaded R-CNN framework in our experiments, as illustrated in Fig. 1. More concretely, the GA modules are connected between the backbone network and FPN, while the TBA module is linked between the FPN and RPN+ROI. Our backbone network of choice is the ResNet50, with the learned features from C={C2, C3, C4, C5} feature maps being utilized. Correspondingly, three GA modules with identical structures are added in between the cross-level combinations of {C2, C3}, {C3, C4}, and {C4, C5}, producing three feature maps F={G2, G3, G4}. These outputs are then provided to FPN along with C5 for further processing.

During the latter inference stage, the TBA module is invoked, whereby the two feature layers P2 and P3 produced by FPN are jointly encoded through Tucker bilinear pooling and self-attention mechanisms to deliver R2, which is ultimately fed into RPN+ROI along with P3, P4, and P5 to complete object detection. It is worth noting that in our experimental setup, only high-resolution feature maps across various layers are passed through TBA to preserve a richer set of salient details.

### B  Guided Attention

Detection networks based on Convolutional Neural Networks (CNN) are capable of efficiently capturing high-level semantic features through convolution and pooling operations. However, this occasionally results in the loss of detailed and intricate aspects associated with smaller objects.

Essentially, shallow features harbor much local detail, whereas deep features predominantly emphasize larger regions, creating insufficient feature discrimination for smaller objects. Against this backdrop, this paper introduces the guided attention (GA) module, which enables more effective mapping between the shallow and deep features. By applying the GA module in our study, the semantic context information in the deep layers is utilized to guide the learning process of the shallow layers. This approach significantly improves the preservation and extraction of relevant features that pertain to small objects.

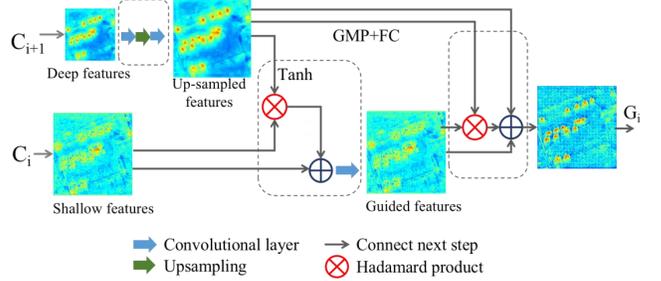

Fig.2. Guided Attention module.

Figure 2 showcases the architectural structure of the guided attention module, which primarily comprises of three operations. Initially, the deeper layer undergoes convolution and up-sampling transformations to generate a feature map with the same dimension as the shallow layer. The corresponding formula is as follows:

$$f_{up} = \text{Conv}_{3 \times 3}((\text{Up}(\text{Conv}_{1 \times 1}(C_{i+1}))) \quad (1)$$

with $C_i \in \mathbb{R}^{\frac{1}{2}C \times 2H \times 2W}$, $C_{i+1} \in \mathbb{R}^{C \times H \times W}$, $f_{up} \in \mathbb{R}^{\frac{1}{2}C \times 2H \times 2W}$. $f_{up}$ is the up-sampled feature map. UP( ) represents an upsampling operation and $\text{Conv}_{3 \times 3}()$ represents a convolutional layer with a 3x3 convolutional kernel.

Proceeding to the second step, we treat the upsampled feature maps as queries and the shallow feature maps as keys, and perform an attention operation. It is noteworthy that instead of employing the typical Softmax activation, we use Tanh activation for this process, yielding guided attention maps as a result. The corresponding formula for this step is as follows:

$$f_{guide} = \text{Conv}_{3 \times 3}(C_i * \text{Tanh}(f_{up}) + C_i) \quad (2)$$

with $f_{guide} \in \mathbb{R}^{\frac{1}{2}C \times 2H \times 2W}$. Tanh() is the tanh activation function.

Finally, we enhance the guided attention map by taking the product with the maximum eigenvalue in the deep layer. We subsequently combine the guided attention map, up-sampled feature map, and enhanced guided attention map through summation. The corresponding formula for this step is outlined below:

$$G_i = W_{fc} * \text{Gmp}(f_{up}) * f_{guide} + f_{guide} + f_{up} \quad (3)$$

with $G_i \in \mathbb{R}^{\frac{1}{2}C \times 2H \times 2W}$. $W_{fc}$ represents the weight of the linear fully connected layer, and Gmp() represents the global maximum pooling.

## C  Tucker Bilinear Attention(TBA)

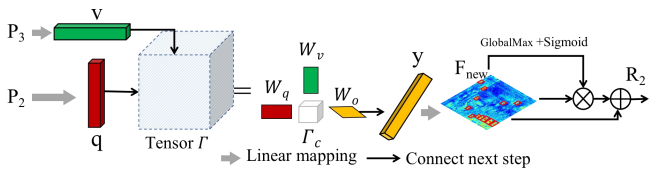

Fig.3. Tucker bilinear attention module

Bilinear models present as an efficient and accomplished solution toward addressing the feature fusion predicament, as they capture a full array of second-order interactions occurring between q and v:

$$y = (\Gamma \times_1 q \times_2 v) \quad (4)$$

with the full tensor $\Gamma \in \mathbb{R}^{I \times J \times K}$, and $\times_n$ indicating the tensor product along the *n*th mode.

Despite their effectiveness, one of the major roadblocks to using bilinear models is their parameter complexity, which escalates rapidly and can surpass manageable thresholds concerning input and output dimensions. Addressing this issue, our study presents a novel approach for feature fusion that seeks to capture intricate and detailed interactions between scales of feature layers. We leverage the Tucker decomposition of the correlation tensor in achieving this objective, which allows us to model complete bilinear interactions while retaining a manageable model size.

The Tucker decomposition [11], is a mathematical technique that decomposes a tensor into a smaller core tensor and a set of associated matrices. Specifically, in the three-mode case, given an original tensor $\Gamma \in \mathbb{R}^{I \times J \times K}$, the Tucker decomposition yields a core tensor $\Gamma_c \in \mathbb{R}^{P \times Q \times R}$ and three matrices $W_q \in \mathbb{R}^{I \times P}$, $W_v \in \mathbb{R}^{J \times Q}$, $W_o \in \mathbb{R}^{K \times R}$.

$$\Gamma \approx \Gamma_c \times_1 W_q \times_2 W_v \times_3 W_o \quad (5)$$

$W_q$, $W_v$ and $W_o$, when orthogonal, can be thought of as the principle components. $\Gamma_c$ shows the level of interaction between the different components. A comprehensive discussion on Tucker decomposition and tensor analysis may be found in [12].

The form of Tucker decomposition is utilized to integrate features from adjacent layers of scales. By adopting Equation 5 as a substitution for Equation 4, we obtain an alternate formulation of the bilinear model, as demonstrated in Equation 6. This model comprises of the q and v variables that serves as inputs to the module, and the $\Gamma_c$, $W_q$, $W_v$ and $W_o$ matrices that are learnable parameters updated through stochastic gradient descent.

$$y = (\Gamma_c \times_1 (q^T W_q) \times_2 (v^T W_v) \times_3 W_o) \quad (6)$$

The process is clearly outlined in Figure 3, whereby the input features undergo matrix multiplication to transform them into low-dimensional vectors. These vectors are referred to as $\tilde{q} = q^T W_q$ and $\tilde{v} = v^T W_v$ after transformation. Subsequently, a mapping latent space is obtained by multiplicatively computing $\tilde{q}$ and $\tilde{v}$ with the tensor core. This step reveals the embedding correlation relationship that exists between different features in the hidden space. Multiplicative transformation is again applied to features in the hidden space to achieve the final output feature y. However, the feature fusion approach that is based on the Tucker decomposition network uses several groups of learnable parameters, which could lead to over-parameterization. To balance the trade-off between complexity and expressiveness of interaction modeling, we propose a structured sparsity constraint that is dependent on the rank of slices in $\Gamma_c$. We discovered that the degree of association between categories has an impact on the accuracy of object detection. For this reason, we restrict the ranks of q and v to be less than or equal to the total number of categories K. Specifically, the dimension of the slice d matrix in $\Gamma_c$ corresponding to the $\tilde{q}$ and $\tilde{v}$ axes is not greater than KxK. In addition to this, we incorporate the L1 norm of $\Gamma_c$ to further constrain its sparsity.

After that, we restore y to the 2D feature map $F_{new}$ through a linear mapping. We added a self-attention structure to further strengthen $F_{new}$. The self attention structure can be expressed by the formula:

$$R_2 = \text{Sigmoid}(\text{Gmp}(F_{new})) * F_{new} + F_{new} \quad (7)$$

with R2 the final output of the TBA module.

## III. EXPERIMENTS AND ANALYSIS

### A. Dataset

To evaluate the efficiency and efficacy of our proposed approach, we conducted a series of experiments on various datasets, including DOTA [13], DIOR [14], and NWPU VHR-10 [15]. In line with the established practice, we ensured that our data partitioning strategy is consistent with the official dataset guidelines.

### B. Evaluation Metrics

We adopt the mAP (mean Average Precision) as a metric to evaluate the accuracy of object detection. This metric is a composite measure calculated by averaging the AP (Average Precision) values for all target categories:

$$mAP = \frac{\sum_i^n AP(x_i)}{K} \quad (8)$$

where AP denotes Average Precision, and K is the number of categories. The integral method is employed for computing the area encompassed by the Precision-Recall curve and the coordinate axis.

### C. Experimental Result

The comparison results with the state-of-the-art methods[16,17,18,19] are shown in TABLE 1, TABLE 2, and TABEL 3 on DOTA, DIOR, and NWPU VHR-10. From the results, our method has greatly improved the detection accuracy of small objects without reducing the detection accuracy of medium and large-sized objects, thus achieving the best mean average precision.

On DOTA, the four best-performing categories of our method are ground field track(C11), storage tank(C1), ship(C3), and small vehicle(C8) with the AP metrics of 72, 88.4, 80.2, 62.4 which are higher than the AP metrics in the second place by 5.6, 5.0, 4.3 and 3.7 percentage points. Among them, C1, C3 and C8 are categories with the largest number of small objects.

On the DIOR dataset, our comprehensive performance is the best. GCF is slightly inferior to ours, and like us, it also attempts to guide and obtain clean features by high-level vectors. On DIOR, the four categories with the highest number of small objects are ships(C14), vehicles (C19), storage tanks (C15), and airplanes(C1). Our method won first place on the AP metrics in all of these categories. Especially in ships and vehicles, the improvement is the most obvious, with AP increased by 7.2 and 1.8 percentage points respectively. Further observations show that our method significantly outperforms other methods in categories containing large objects, such as harbors and basketball courts.

On NWPU VHR10, similar phenomena can be observed, that is, our method has the best mean average precision, and performs best on categories with more small targets, as shown in Table 3. Except that our method is inferior to some individual methods in the categories of the ground track field (C7) and harbor(C8), it won first place in all the other categories.

### F. Ablation Study

TABLE 5 Ablation studies

| Model | | mAP | C1 | C2 | C3 |
|---|---|---|---|---|---|
| GA | TBA | | C11 | C12 | C13 |
| | | 70.0 | 62.7 | 85.6 | 72.0 |
| | | | 82.1 | 57.1 | 63.4 |
| √ | | 72.4 | 72.3 | 87.0 | 72.2 |
| | | | 80.2 | 65.9 | 63.1 |
| | √ | 72.1 | 63.2 | 87.3 | 72.1 |
| | | | 81.9 | 65.6 | 63.7 |
| √ | √ | 73.3 | 72.3 | 87.5 | 72.3 |
| | | | 81.9 | 65.9 | 63.9 |

To further understand the effectiveness of the proposed method, we conduct ablation studies. We choose Cascaded R-CNN as the baseline, and try to experiment in the following four cases: 1) baseline, 2) baseline plus the TBA module, 3) baseline plus the GA modules, and 4) our method with both TBA and GA. We carried out ablation experiments on the DOTA, DIOR, and NWPU VHR-10 datasets and the results are shown in TABLE 4.

From the results of ablation experiments, it can be observed that although adding any module alone can improve the accuracy of object detection, the combination of the two works best. FA helps to preserve more detail features effectively, while TBA further improves attention on relevant detail features before inference.

### G. Qualitative Analysis

Fig.4, Fig.5 show the qualitative analysis results for representative samples. As can be seen from Fig.10, the results of our method are very close to the ground truth and vehicles of different sizes can be detected accurately. Other comparison methods only perform well on relatively large vehicles, existing the phenomenon of missed detection of small vehicles.

Figure 11 shows the situation of densely packed small targets. In this case, it is easy to miss small objects and mistakenly detect large objects. The compared methods have different degrees of false detection and missed detection, while the results of our method are exactly the same as the ground truth. Objects of different categories may vary greatly in scale. According to qualitative analysis, we can draw the same conclusion as before, that is, the proposed method can significantly reduce the missed and false detection of small objects without degradation of medium and large-sized objects, thus improving mAP on all the dataset.

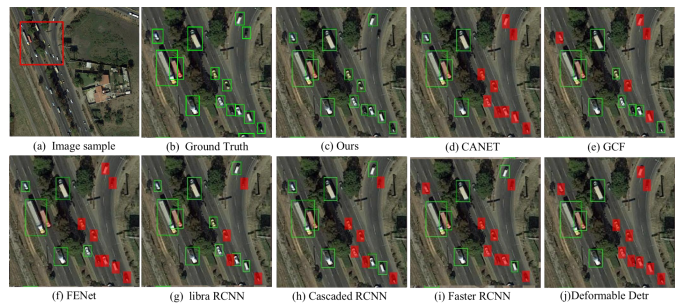

(a) Image sample (b) Ground Truth (c) Ours (d) CANET (e) GCF
(f) FENet (g) libra RCNN (h) Cascaded RCNN (i) Faster RCNN (j)Deformable Detr

Fig.10 Comparison results on a sample image which contains vehicles of different sizes. Red boxes denote the false or missed detection.

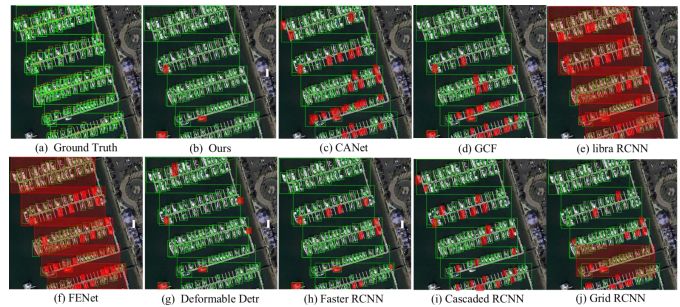

(a) Ground Truth (b) Ours (c) CANet (d) GCF (e) libra RCNN
(f) FENet (g) Deformable Detr (h) Faster RCNN (i) Cascaded RCNN (j) Grid RCNN

Fig.11 Comparison results on a sample image which contains harbors and densely packed ships. Red boxes denote the false or missed detection.

### IV. CONCLUSION

This paper proposes an object detection framework based on Tucker bilinear attention to improve the detection accuracy of multi-scale objects in remote sensing imagery, especially small objects. We suggest two novel modules: a Guided Attention (GA) module and a Tucker Bilinear Attention(TBA) module. The former guides shallow features through deep features to better retain details; the latter fuses and balances the features by constructing dynamically learnable sparse matrices. Comparing experiments with 9 state-of-the-art methods on three datasets show that the proposed method achieves the best mAP for objects of different scales.

TABLE 1 The results compared with the SOTA methods on the DOTA dataset

| Model | mAP | C1 | C2 | C3 | C4 | C5 | C6 | C7 | C8 | C9 | C10 | C11 | C12 | C13 | C14 | C15 |
|---|---|---|---|---|---|---|---|---|---|---|---|---|---|---|---|---|
| Faster R-CNN | 69.0 | 80.7 | 78.0 | 69.9 | 52.1 | 59.3 | 54.9 | 89.3 | 53.8 | 78.2 | 74.9 | 65.1 | 65.7 | 90.7 | 69.1 | 53.8 |
| Cascade R-CNN | 69.2 | 82.7 | 77.6 | 75.9 | 49.8 | 59.4 | 54.0 | 89.2 | 57.4 | 77.1 | 74.1 | 65.9 | 62.9 | 90.8 | 67.4 | 53.6 |
| Libra R-CNN | 69.2 | 80.4 | 77.4 | 69.7 | 50.9 | 63.9 | 61.5 | 89.2 | 52.5 | 77.8 | 73.8 | 65.7 | 65.6 | 90.7 | 68.5 | 51.1 |
| Grid R-CNN | 70.3 | 85.9 | 78.1 | 70.1 | 52.3 | 63.0 | 67.7 | 89.1 | 52.6 | 75.5 | 76.9 | 65.4 | 68.0 | 90.8 | 67.4 | 52.3 |
| Dynamic R-CNN | 68.1 | 80.3 | 75.8 | 69.6 | 49.6 | 59.3 | 61.6 | 89.3 | 50.3 | 77.4 | 73.5 | 63.5 | 63.0 | 90.8 | 69.6 | 46.6 |
| Deformable Detr | 63.4 | 77.5 | 69.2 | 66.7 | 49.6 | 55.4 | 53.4 | 88.8 | 50.3 | 76.0 | 68.9 | 62.9 | 64.7 | 65.5 | 57.2 | 44.0 |
| FENet | 67.4 | 80.5 | 76.8 | 70.7 | 49.8 | 61.1 | 55.9 | 88.8 | 54.1 | 78.1 | 71.3 | 62.3 | 64.5 | 90.6 | 61.5 | 46.3 |
| GCF-Net | 70.3 | 84.7 | 83.3 | 67.8 | 52.1 | 65.6 | 57.6 | 88.8 | 48.5 | 77.5 | 79.9 | 66.4 | 68.5 | 90.7 | 64.7 | 43.7 |
| CANet | 70.2 | 80.5 | 77.7 | 69.2 | 46.9 | 59.7 | 63.5 | 88.5 | 53.4 | 77.3 | 71.1 | 60.5 | 66.7 | 90.3 | 62.7 | 40.3 |
| ours | 71.5 | 88.4 | 78.7 | 80.2 | 52.9 | 61.6 | 57.0 | 90.1 | 62.4 | 78.6 | 73.6 | 72.0 | 64.0 | 90.9 | 68.1 | 54.7 |

TABLE 2 The results compared with the SOTA methods on the DIOR dataset

| Model | mAP | C1 | C2 | C3 | C4 | C5 | C6 | C7 | C8 | C9 | C10 |
|---|---|---|---|---|---|---|---|---|---|---|---|
|       |     | C11 | C12 | C13 | C14 | C15 | C16 | C17 | C18 | C19 | C20 |
| Faster R-CNN | 69.2 | 62.9 | 83.9 | 71.6 | 88.2 | 50.0 | 72.6 | 67.6 | 83.2 | 70.4 | 76.5 |
|              |      | 82.1 | 54.4 | 63.1 | 72.0 | 72.3 | 62.5 | 81.5 | 65.4 | 43.3 | 81.3 |
| Cascade R-CNN | 70.0 | 62.7 | 85.6 | 72.0 | 87.3 | 50.4 | 72.5 | 62.2 | 79.1 | 70.8 | 76.4 |
|               |      | 82.1 | 57.1 | 63.4 | 72.1 | 75.1 | 62.5 | 81.5 | 62.5 | 43.5 | 80.9 |
| Libra R-CNN | 70.3 | 70.8 | 81.9 | 77.8 | 87.6 | 48.9 | 72.5 | 66.4 | 80.6 | 72.2 | 76.0 |
|             |      | 81.8 | 52.0 | 60.3 | 71.9 | 76.2 | 60.6 | 81.3 | 62.7 | 42.4 | 81.3 |
| Grid R-CNN | 71.1 | 62.2 | 83.5 | 76.9 | 87.1 | 49.3 | 72.2 | 64.4 | 80.4 | 69.7 | 78.9 |
|            |      | 80.9 | 57.7 | 60.2 | 72.0 | 75.0 | 62.5 | 81.4 | 63.4 | 43.1 | 81.2 |
| Dynamic R-CNN | 69.3 | 54.2 | 82.0 | 71.7 | 87.4 | 48.4 | 72.5 | 67.9 | 78.3 | 70.6 | 75.7 |
|               |      | 82.9 | 52.4 | 61.8 | 71.9 | 75.9 | 62.1 | 81.5 | 64.5 | 42.8 | 81.2 |
| Deformable Detr | 69.3 | 68.0 | 82.4 | 78.8 | 88.1 | 42.9 | 72.5 | 72.3 | 81.8 | 72.3 | 78.0 |
|                 |      | 79.3 | 48.1 | 60.9 | 71.5 | 63.2 | 59.0 | 82.0 | 60.9 | 39.8 | 83.4 |
| FENet | 68.3 | 54.1 | 78.2 | 71.6 | 81.0 | 46.5 | 79.0 | 65.2 | 76.5 | 69.6 | 79.1 |
|       |      | 82.2 | 52.0 | 57.6 | 71.9 | 71.8 | 62.3 | 81.2 | 61.2 | 43.3 | 68.3 |
| GCF-Net | 72.6 | 68.0 | 87.0 | 74.9 | 88.9 | 47.8 | 77.7 | 68.8 | 84.2 | 71.3 | 76.9 |
|         |      | 83.1 | 59.0 | 61.3 | 73.6 | 76.2 | 62.1 | 87.6 | 67.8 | 46.7 | 88.6 |
| CANet | 71.5 | 72.2 | 84.3 | 72.1 | 89.0 | 52.4 | 72.2 | 69.5 | 83.1 | 71.1 | 77.3 |
|       |      | 82.5 | 63.1 | 63.4 | 72.1 | 74.5 | 62.3 | 81.5 | 67.3 | 43.7 | 81.4 |
| ours | 73.3 | 72.3 | 87.5 | 72.3 | 89.0 | 53.7 | 72.5 | 71.0 | 85.1 | 77.6 | 78.1 |
|      |      | 81.9 | 65.9 | 63.9 | 80.8 | 76.2 | 62.5 | 81.5 | 65.5 | 48.5 | 80.9 |

TABLE 3 The results compared with the SOTA methods on the NWPU-VHR10 dataset

| Model | mAP | C1 | C2 | C3 | C4 | C5 | C6 | C7 | C8 | C9 | C10 |
|---|---|---|---|---|---|---|---|---|---|---|---|
| Faster R-CNN | 81.6 | 100.0 | 45.7 | 90.6 | 89.1 | 80.2 | 89.1 | 89.1 | 84.6 | 88.8 | 59.5 |
| Cascade R-CNN | 82.3 | 99.9 | 58.1 | 90.4 | 90.5 | 68.0 | 96.1 | 89.3 | 78.6 | 89.2 | 62.8 |
| Libra R-CNN | 83.1 | 99.9 | 46.9 | 90.5 | 89.2 | 79.0 | 90.3 | 88.5 | 92.8 | 90.0 | 63.6 |
| Grid R-CNN | 83.9 | 100.0 | 58.7 | 90.8 | 89.6 | 76.8 | 97.0 | 89.6 | 84.7 | 88.3 | 63.6 |
| Dynamic R-CNN | 80.4 | 90.9 | 48.8 | 90.5 | 90.7 | 83.8 | 87.0 | 89.0 | 79.2 | 90.0 | 54.5 |
| FENet | 82.9 | 99.9 | 58.4 | 57.1 | 90.6 | 90.2 | 95.2 | 98.7 | 87.3 | 81.9 | 69.7 |
| GCF-Net | 84.0 | 100.0 | 56.2 | 90.8 | 90.6 | 79.5 | 99.4 | 89.2 | 80.3 | 90.2 | 63.6 |
| CANet | 83.1 | 99.9 | 54.2 | 90.8 | 90.7 | 81.3 | 99.1 | 89.5 | 90.0 | 81.8 | 54.5 |
| ours | 90.1 | 100.0 | 69.9 | 99.4 | 90.7 | 90.5 | 99.4 | 90.0 | 90.3 | 90.7 | 80.0 |